\documentclass[sigconf]{acmart} 

\renewcommand\footnotetextcopyrightpermission[1]{} 
\AtBeginDocument{%
  \providecommand\BibTeX{{%
    \normalfont B\kern-0.5em{\scshape i\kern-0.25em b}\kern-0.8em\TeX}}}

\setcopyright{acmcopyright}
\copyrightyear{2023}
\acmYear{2023}
\acmDOI{xx.xxxx/xxxxxxx.xxxxxxx}

\acmConference[Under Review]{Under Review}
\acmBooktitle{}

\acmPrice{15.00}
\acmISBN{978-1-4503-XXXX-X/18/06}

\usepackage{algorithm}
\usepackage{algorithmicx}
\usepackage{algpseudocode}
\usepackage{amsmath, bm}
\usepackage{subfig}
\usepackage{booktabs}
\usepackage{multirow}
\usepackage{graphicx}
\usepackage{enumitem}
\usepackage[thicklines]{cancel}
\usepackage{bbding}
\usepackage{tabularx}
\usepackage[linewidth=1pt]{mdframed}

\begin{document}

\title{Beyond Static Datasets: A Deep Interaction Approach to LLM Evaluation}
\renewcommand{\shorttitle}{}


\author{Jiatong Li}
\authornote{Both authors contributed equally to this research.}
\email{satosasara@mail.ustc.edu.cn}
\affiliation{%
  \institution{School of Data Science, University of Science and Technology of China}
  \city{Hefei}
  \state{Anhui Province}
  \country{China}
  \postcode{230031}
}

\author{Rui Li}
\authornotemark[1]
\email{ruili2000@mail.ustc.edu.cn}
\affiliation{%
  \institution{School of Computer Science, University of Science and Technology of China}
  \city{Hefei}
  \state{Anhui Province}
  \country{China}
  \postcode{230031}
}

\author{Qi Liu}
\email{qiliuql@ustc.edu.cn}
\affiliation{%
  \institution{School of Computer Science, University of Science and Technology of China}
  \city{Hefei}
  \state{Anhui Province}
  \country{China}
  \postcode{230031}
}

\renewcommand{\shortauthors}{}

\begin{abstract}
  Large Language Models (LLMs) have made progress in various real-world tasks, which stimulates requirements for the evaluation of LLMs. Existing LLM evaluation methods are mainly supervised signal-based which depends on static datasets and cannot evaluate the ability of LLMs in dynamic real-world scenarios where deep interaction widely exists. Other LLM evaluation methods are human-based which are costly and time-consuming and are incapable of large-scale evaluation of LLMs. To address the issues above, we propose a novel Deep Interaction-based LLM-evaluation framework. In our proposed framework, LLMs' performances in real-world domains can be evaluated from their deep interaction with other LLMs in elaborately designed evaluation tasks. Furthermore, our proposed framework is a general evaluation method that can be applied to a host of real-world tasks such as machine translation and code generation. We demonstrate the effectiveness of our proposed method through extensive experiments on four elaborately designed evaluation tasks. Our source code is available at \url{https://anonymous.4open.science/r/DeepEval-112F}.
\end{abstract}





\maketitle
  
  \section{Introduction}
  \par With the rapid growth of Large Language Models (LLMs), LLM-based applications have made progress and exceeded human performance in many real-world domains such as machine translation and code generation. The advancement of LLM-based applications also facilitates the requirement for the evaluation of LLMs. Due to the huge scale and the lack of interpretability of LLMs, the evaluation of LLMs mainly focuses on their skill sets on domain-specific tasks. The evaluation results of LLMs later can guide users to choose appropriate LLMs for their unique requirements.

  \begin{figure}[t]
      \centering
      \includegraphics[width=\linewidth]{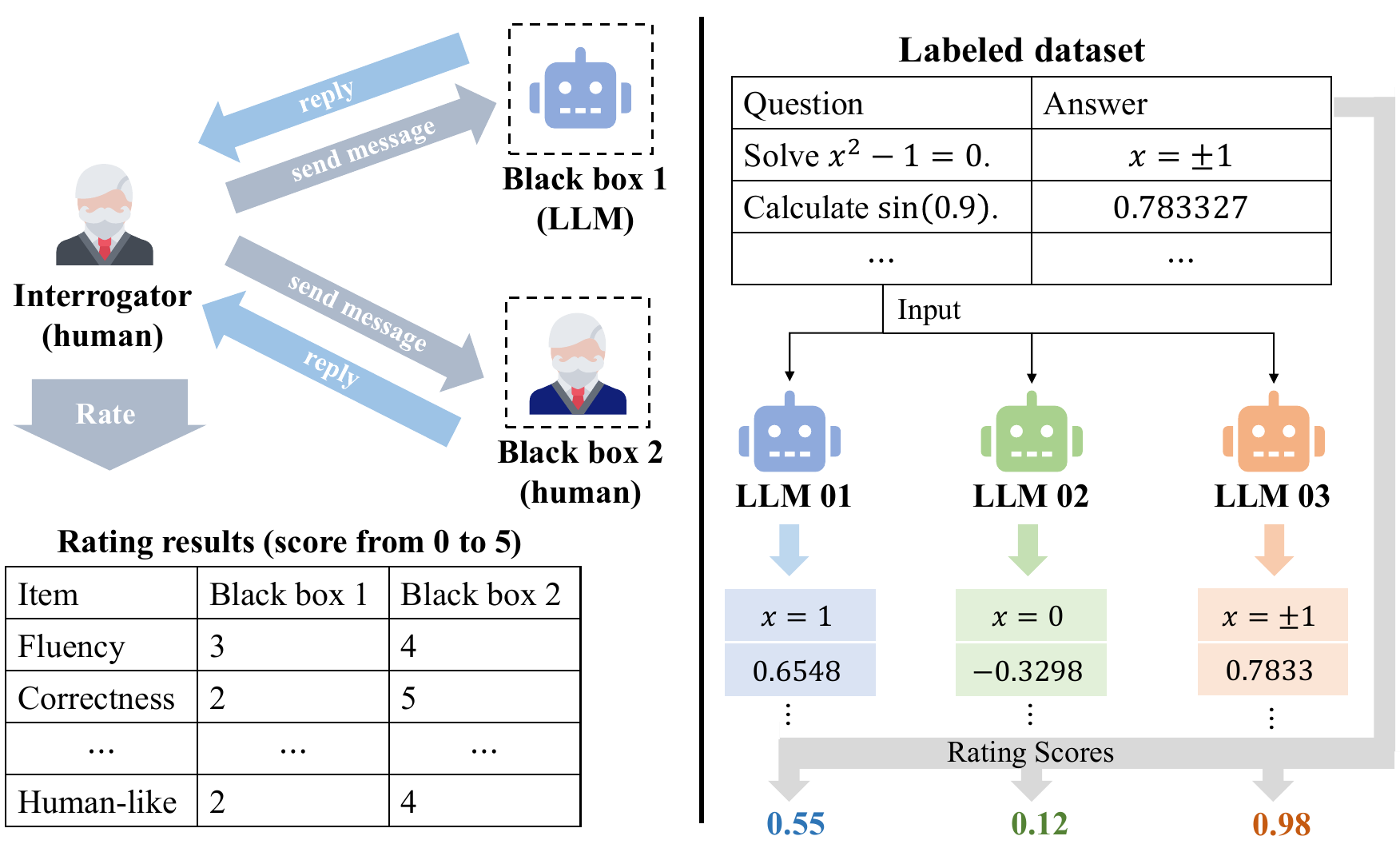}
      \caption{\textbf{Left}: Turing test, an example of human-based evaluation methods. \textbf{Right}: mathematical question answering, an example of supervised signal-based evaluation methods}
      \label{fig:turing-test}
      \vspace{-15pt}
  \end{figure}
  
  \par In the literature, traditional LLM-evaluation methods are either human-based or supervised signal-based, as shown in Figure \ref{fig:turing-test}. In human-based methods, human interrogators interact with LLMs, and the evaluation of LLMs depends on the judgment of human interrogators. For example, in the Turing test \cite{shieber2006turing}, a human interrogator can interact with two anonymous examinees (one is an LLM while the other is a human) and is required to distinguish the LLM from the human in a limited time. Despite their flexibility, human-based evaluation methods are too costly to be applied to the large-scale evaluation of LLMs in numerous tasks. In supervised signal-based evaluation methods, LLMs are required to generate correct outputs given dataset inputs. Compared to human-based evaluations, supervised signal-based evaluations are more automatic because evaluation results of LLMs can be gained by automatically calculating evaluation metrics based on dataset labels and LLM outputs. Thus supervised signal-based methods have been widely applied to large-scale evaluations of LLMs. However, there are two shortcomings in these methods. First, datasets in supervised signal-based evaluations are static, which makes it impossible for these methods to evaluate the deep interaction ability of LLMs. In real-world scenarios, LLMs often interact with users in multiple rounds to receive feedback and fix their outputs to meet the users' needs. As a result, the static supervised signal-based methods cannot reflect the true performance of LLMs in these scenarios. Second, LLMs often act in different roles even in the same task. For example, in code generation, some users need LLMs to act as programmers to generate complete codes, while others need LLMs to act as reviewers to optimize their own codes. However, in supervised signal-based methods, LLMs can only act one role in a dataset. This fact limits the range of evaluated skills, and the ability of the method to discover the potential of LLMs to act in different roles.
  \par To address the issues above, in this paper, we propose a novel deep interaction-based LLM-Evaluation Framework. Our motivation is that deep interactions between LLMs and users can be simulated by the interaction of various LLMs in elaborately designed evaluation tasks, and LLMs' deep interaction abilities and domain-specific skill proficiencies can be evaluated from their interaction records. Specifically, we first propose a general evaluation framework that can be integrated with various evaluation tasks. Next, we propose a general evaluation running algorithm to ensure the correctness and fairness of the evaluation process. Furthermore, we analyze and give the principle of the design of deep interaction-based evaluation tasks. In experiments, we evaluate the performance of four well-known LLMs with our proposed framework in four elaborately designed evaluation tasks, i.e., public goods game, idioms solitaire, machine translation, and code generation.
  \par In a word, our contribution can be summarized as follows:
  \begin{itemize}[leftmargin=*]
      \item We propose a deep interaction-based evaluation framework for evaluating LLMs in real-world scenarios with deep interactions, which can be integrated with a host of tasks.
      \item We analyse and present the principle of the design of evaluation tasks, and give four examples for experiments.
      \item We demonstrate the effectiveness of our proposed method by extensive experiments on four elaborately designed tasks that evaluate both the deep interaction abilities and domain-specific skills of four well-known LLMs.
  \end{itemize}
  
  \section{Related Works}
  \subsection{The Development of LLMs}
  \par Large language models (LLMs) are sophisticated artificial intelligence systems that are designed to understand and generate human-like text to aid humans in all kinds of real-world tasks. LLMs are trained on vast amounts of data and utilize deep learning techniques to learn patterns, language structures, and relationships between words and sentences. So far, LLMs have revolutionized the field of natural language processing (NLP) and have been the subject of extensive research and development. These models, such as BERT (Bidirectional Encoder Representation from Transformers) \cite{vaswani2017attention} and GPT-3 (Generative Pre-trained Transformer 3), have demonstrated impressive capabilities in various NLP tasks, including text generation, machine translation, question answering, summarization, etc. 
  \par Recently, starting from the presence of ChatGPT\footnote{\url{https://openai.com/chatgpt}}, more and more LLMs are proposed to solve various real-world challenges. For example, Claude\footnote{\url{https://www.anthropic.com/claude-in-slack}} is an LLM application proposed by Antropic, which is good at many real-world tasks such as paper writing and coding. PaLM \cite{chowdhery2022palm} is a scaling LLM proposed by Google, and shows strong capabilities in multilingual tasks and code generation. Indeed, the recent advancement of LLMs has shown their power in boosting people's productivity. However, despite their capabilities, LLMs are black boxes that lack explainability, and we cannot ensure their competitive performance in any real-world situation. As a result, researchers manage to evaluate LLMs in various scenarios to quantitatively measure their capabilities in different tasks.
  
  \subsection{The Evaluation of LLMs}
  \par The recent advancement of LLMs in real-world scenarios stimulate the requirement for the evaluation of LLMs \cite{changsurvey2023}. In the literature, the evaluation of LLMs can be human-based or supervised signal-based. Specifically, human-based evaluation depends on human interrogators to measure the performance of LLMs. For example, Likert scale \cite{petrillo2011likert} is a representative method that utilizes a rating scale filled by human judgements to measure the performance of LLMs in different dimensions (e.g., fluency, coherence). Despite their flexibility, human-based evaluations are costly and time-consuming, thus they incapable for large-scale evaluation of LLMs in a host of real-world tasks. On the other hand, supervised signal-based evaluation depends on supervised signals in expert-labelled datasets to evaluate the performance of LLMs. Supervised signal-based evaluations are efficient, and have been applied to large-scale evaluation of LLMs in many fields such as machine translation\cite{bang2023multitask, lyu2023new, wang2023documentlevel}, reasoning\cite{frieder2023mathematical, saparov2023testing,orru2023humanlike} and code generation\cite{kashefi2023chatgpt,zhuang2023efficiently,hendrycks2021apps}. Despite their efficiency, supervised signal-based evaluation of LLMs depends on static datasets and cannot evaluate the ability of LLMs in dynamic real-world scenarios where deep interaction between LLMs and users widely exists, thus the evaluation correctness is limited.
  \par Recently, there are also some studies committed to more automated and general evaluation of LLMs. For instance, MT-Bench \cite{zheng2023judging} utilizes an LLM as a judgement to automatically measure the performance of LLMs in multi-round conversations. However, conversations scenarios in this method are fixed limited to the static dataset. PandaLM \cite{wang2023pandalm} further utilizes a well-designed discriminative large-scale language model to evaluate the performance of LLMs in different tasks, and can provide more fine-grained evaluation results in terms of subjective elements compared to conventional evaluation methods. AlpacaEval \cite{dubois2023alpacafarm} also focuses on the autometic evaluation of LLMs. As an automated evaluation benchmark, AlpacaEval can evaluate the performance of LLMs in a host of natural language processing tasks. GameEval \cite{qiao2023gameeval} utilizes conversational games to gain more distinguishable evaluation results of LLMs. PromptBench \cite{zhu2023promptbench} focuses on the adversarial prompt resilience, and is able to evaluate the adversarial robustness of LLMs. Although these evaluation methods have shown generality to some extent, they are either supervised signal-based or limited by scenarios. As a result, these evaluation methods are still hard to evaluate LLMs in complex real-world tasks where deep interaction widely exists.
  
  \section{Our Proposed Framework}
  \subsection{Preliminaries}
  \par In this part, we first introduce mathematical notations. Next, we give the formal definition of deep interaction-based tasks for evaluation. Finally, we give the formal definition of deep interaction-based evaluation of LLMs.
  \par To begin with, let $N$ denote the number of LLMs to be evaluated. Let $\mathcal{P}=\{P_1,P_2,\ldots,P_N\}$ denote the set of LLMs where each element denotes an LLM. As for evaluation tasks, inspired by game theories\cite{kreps1982games, apt2021extensive}, we leverage the properties of extensive games with perfect information to define deep interaction-based evaluation tasks. The core of extensive games with perfect information is that participants can consider their plan with perfect information at any time when it has to make a decision, which is consistent with real-world deep interaction scenarios where LLMs have to make a decision based on all of their historical interactions with users. To this end, let $R$ be a deep interaction-based evaluation task, then it is defined as follows:

  \begin{definition}\label{def:game}
      \textbf{Deep interaction-based evaluation tasks}. a deep interaction-based evaluation task is defined a collection of component listed as follows:
      \begin{itemize}[leftmargin=*]
          \item A set $\mathcal{P}=\{P_1,P_2,\ldots,P_N\}$ (the set of $N$ LLMs).
          \item A history set $H$ of sequences (finite or infinite), which can be represented by a gaming tree. $H$ satisfies the following properties:
          \begin{enumerate}
              \item The empty sequence $\emptyset$ is a member of $H$, i.e., $\emptyset\in H$, which serves as the root node of the gaming tree.
              \item If a sequence $(a_k)_{k=1,\ldots,K}\in H$ and $L < K$, then $(a_k)_{k=1,\ldots,L}\in H$. Further, if $(a_k)_{k=1,\ldots,K+1}\notin H$, then $(a_k)_{k=1,\ldots,K}$ is a \textbf{terminal} history. 
              \item If an infinite sequence $(a_k)_{k=1}^\infty$ satisfies $(a_k)_{k=1,\ldots,L}\in H$ for every positive integer $L$, then $(a_k)_{k=1}^\infty\in H$.
          \end{enumerate}
          \item A terminal history set $Z$ consisting of all terminal histories, i.e., $Z = \{(a_k)_{k=1,\ldots,K} | (a_k)_{k=1,\ldots,K}\in H, (a_k)_{k=1,\ldots,K+1}\notin H\}$.
          \item An LLM function $P_f: H\backslash Z\to Q$ that assigns to each non-terminal history $h\in H\backslash Z$ a member of LLMs ($P_f(h)$ is the next LLM making decision given the history sequence $h$).
          \item A payoff function $u_i: Z\to\mathbb{R}$ for every LLM $Q_i\in Q$ ($u_i(z)$ is the payoff of LLM $Q_i$ given the terminal history $z\in Z$). 
      \end{itemize}
  Then a deep interaction-based evaluation task can be represented as a tuple, i.e., $R = \langle \mathcal{P},H,Z,P_f,U\rangle$, where $U=\{u_1,\ldots,u_N\}$ is the set of payoff functions.
  \end{definition}
  \par Given the definition of evaluation tasks which can be represented by texts and input to LLMs, LLMs can be asked to interact with other participants to maximize their payoffs. Then the abilities of LLMs can be evaluated from their interaction histories. Specifically, let $\Theta = ({\theta}_1,{\theta}_2\ldots,{\theta}_N)$ be the ability of target LLMs, then LLM evaluation task can be defined as follows:
  
  \begin{definition} \label{def:task}
      \textbf{Deep interaction-based evaluation of LLMs}. Given a set of LLMs $\mathcal{P}$ and an evaluation task $R$, the goal of the deep interaction-based evaluation of LLMs is to evaluate LLMs' abilities $\Theta$ from observed history sequences of LLMs in the game.
  \end{definition}
  
  \subsection{Necessary Conditions of DeepEval}
  \newtheorem{condition}{Condition}
  \par The top priority of deep interaction-based evaluation methods is to ensure the correctness of the evaluation. Thus it is necessary to ensure the fairness of the evaluation process and stableness of evaluation results. To this end, we propose the fairness condition and the stableness condition of DeepEval, which constitute the prerequisite of the design of the framework.
  \begin{condition}
      \textbf{Fairness condition}. All of LLMs that participate in the interaction process should be anonymous, and the delivery of LLMs' messages should be synchronous.
  \end{condition}
  \par The fairness condition essentially ensures the fairness of the evaluation process from two aspects. First, the anonymity of LLMs avoids biased interaction policies of LLMs. For example, if one LLM acknowledges the identity of another LLM, then it can use strategies targeted to the flaw of the latter to win the evaluation task. Second, the synchronicity of the delivery of LLMs' messages guarantees that each LLM has the equal chance to collect information and make decisions.
  \par Next, we introduce the stableness condition which aims to obliterate the influence of uncertainties on evaluation results. The stableness condition is defined as following.
  \begin{condition}
      \textbf{Stableness condition}. To evaluate a set of LLMs, the deep interaction-based evaluation process should be run independently for multiple times until evaluation results of LLMs converge.
  \end{condition}

  \begin{figure*}[t]
      \centering
      \includegraphics[width=\linewidth]{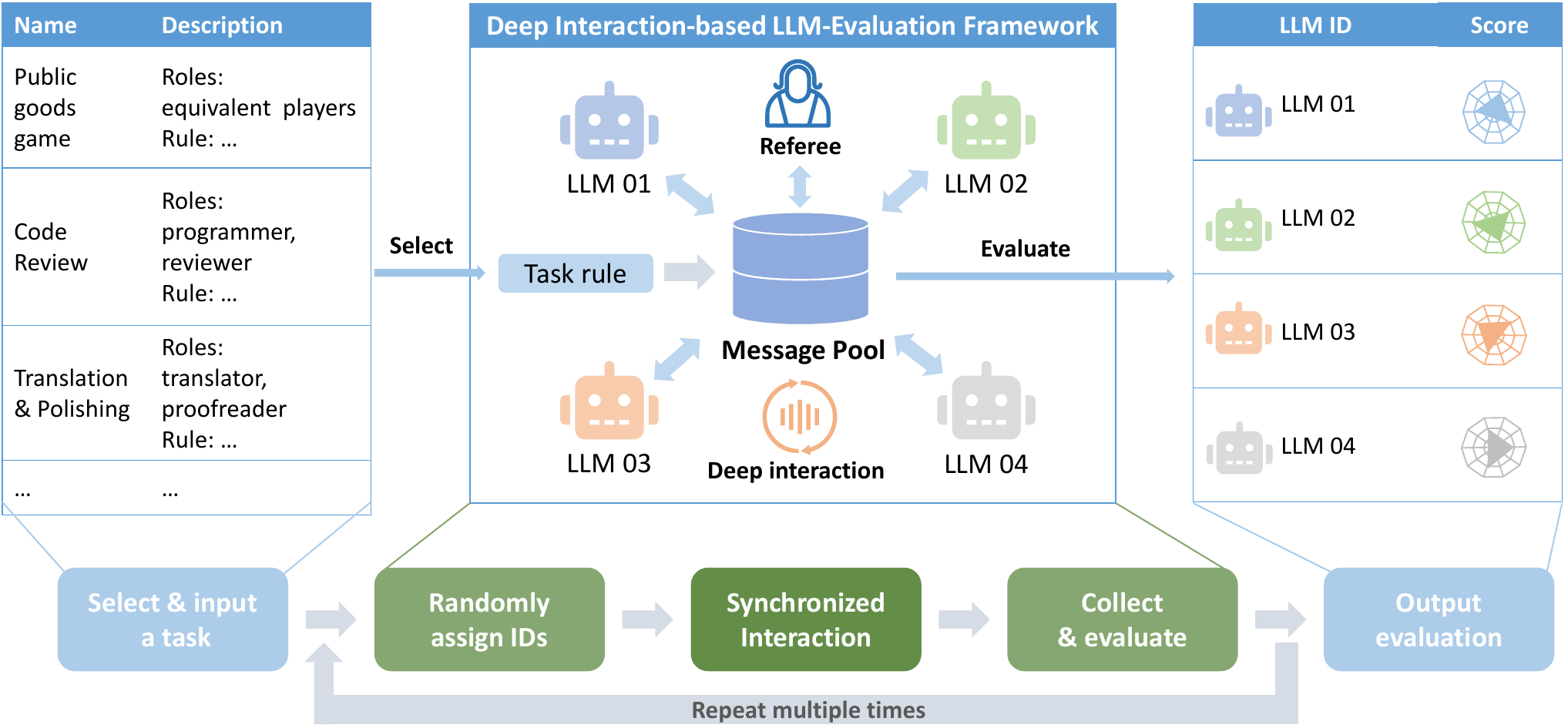}
      \caption{The deep interaction-based LLM evaluation framework.}
      \label{fig:framework}
  \end{figure*}
  
  \subsection{The Structure of DeepEval}
  \par The structure of the deep interaction-based LLM evaluation framework (DeepEval) is shown in Figure \ref{fig:framework}. The motivation of the design of DeepEval is to ensure the fairness of the evaluation process through well-designed mechanisms of the framework. To this end, we propose two pivotal components of DeepEval, i.e., the message pool and the referee. We first introduce the message pool and the referee respectively, then we describe the workflow of DeepEval to illustrate how they work together to ensure the fairness of the evaluation process.
  \begin{itemize}[leftmargin = *]
      \item \textbf{Message Pool}. The message pool in DeepEval is used for containing the interaction histories of LLMs, which serves as the history set $H$ in our formal definition of deep interaction-based evaluation tasks. The message pool is managed by the referee who can read messages and write messages. Indeed, the message pool blocks direct interaction of LLMs, which can ensure the synchronicity of the interaction process together with the synchronous interaction algorithm, which will be introduced later.
      \item \textbf{Referee}. The referee in DeepEval is a pivotal role that is responsible for supervising the evaluation task process and evaluating the performance of LLMs based on interaction histories. Inspired by studies in LLM-as-a-judge \cite{wang2023pandalm,zheng2023judging}, the referee in DeepEval is acted by an LLM.
  \end{itemize}

  \par \textbf{The workflow of DeepEval.} DeepEval starts with an evaluation task box, which stores the information of evaluation tasks. To evaluate LLMs, DeepEval first select an evaluation task from the evaluation task box and input the formalized rule to the message pool, as shown in the middle part of Figure \ref{fig:framework}. Then each LLM in the DeepEval is assigned a unique ID by the referee (either LLM or human). Next, the referee starts and monitors the deep interaction process. In the deep interaction process, the interaction of LLMs is done through the public message pool, which is used for ensuring the synchronicity of communication. As the referee judges that the task comes to an end, it collects and evaluates the performance of LLMs according to their interaction records. The running of the evaluation task is repeated for multiple times to ensure the stableness condition. For the fairness condition, DeepEval satisfies the condition from two aspects. First, for anonymity, LLMs are assigned identity-irrelevant IDs in DeepEval. Second, for synchronicity, we propose a synchronous interaction algorithm to ensure the synchronicity of the interaction process, which will be introduced in the next part. Finally, After finishing the evaluation task, scores of LLMs are utilized to calculate the final evaluation results.
  
  \subsection{Deep Interaction Algorithm}
  \par In DeepEval, LLMs interact with each other to collect information that helps make a decision. To satisfy the fairness condition of DeepEval, we propose a synchronous interaction algorithm that ensures the synchronicity of the interaction process of LLMs, as shown in Algorithm \ref{alg:algorithm}. The basic idea is that \textit{1.} the entire interaction process of LLMs can be decomposed into the interaction process on \textit{rounds}, and \textit{2.} the communication of LLMs can be done by a public message pool managed by the referee (see Figure \ref{fig:framework}) which can block LLMs to send messages until all LLMs have received messages from the last round. Specifically, at the beginning of each round, each LLM sends its reply of its contexts (which is equivalent to the non-terminal history sequence $h\in H\backslash Z$ of the game $R=\langle \mathcal{P},H,Z,P_f,U\rangle$) to the public message pool. Next, for each LLM that can receive messages from others according to the task rule, the referee selects messages from the message pool, and sends these messages to the LLM. At the end of each round, the referee judges whether the game status has come to an end. 
  
  \begin{algorithm}[t]
  \caption{Synchronous Interaction}
  \label{alg:algorithm}
  \begin{algorithmic}[1] 
  \Statex \textbf{Input}: Evaluation task rule $R$, the set of LLM participants $\mathcal{P}$ 
  \Statex \textbf{Output}: Evaluation results $\Theta$

  \State Initialize referee $P_{0}$
  \State Initialize LLMs $P_1, P_2, \ldots, P_N\in\mathcal{P}$
  \State Initialize message pool $B_{msg}$
  \For{$i$ in $1,2,\ldots,N$}
  \State $P_{0}$.send($P_i$, $R$.text)      \Comment{Broadcasting the task rule to LLMs.}
  \EndFor
  
  \For{$i_{round}$ in $1,2,\ldots,M$}
  \For{$j$ in $R$.sendRoleSet($i_{round}$)}
  \State $s_{msg}\leftarrow P_{0}$.getMessage($P_j$)
  \State $B_{msg}$.add($s_{msg}$)           \Comment{LLMs send messages.}
  \EndFor
  \For{$j$ in $R$.receiveRoleSet($i_{round}$)}
  \State $s_{msg}\leftarrow$ $P_0$.select($B_{msg}$, $P_j$, $R$) 
  \State $P_0$.send($P_j$, $s_{msg}$)       \Comment{LLMs receive messages.}
  \EndFor
  \If{$P_{0}$.judgeEnd($R$,$B_{msg}$) is $True$}
  \State Break                              \Comment{The referee judges whether the task is ended.}
  \EndIf
  \EndFor
  \State Let $\Theta\leftarrow P_{0}$.evaluate($B_{msg}$)
  \State \textbf{return} $\Theta$
  \end{algorithmic}
  \end{algorithm}

  \subsection{How to Design Evaluation Tasks}
  \par The design of evaluation tasks in DeepEval is significant because it decides what to evaluate and how to evaluate. For the first aspect, the design of an evaluation task should be consistent with real-world tasks and require relevant skills such as machine translation and code review. For the second aspect, the rule of the evaluation task regularizes the interaction of LLMs, thus defines how to evaluate. To this end, inspired by game thories, we propose the symmetric design and asymmetric design of evaluation tasks, which evaluate LLMs from different perspective.
  \begin{itemize}[leftmargin=*]
      \item \textbf{Symmetric evaluation task}. In symmetric evaluation tasks, all LLMs act the same role with the same action set and task goals. Because task goals of LLMs are the same, this type of evaluation task can evaluate domain-specific skills of LLMs in a competitive manner. Symmetric evaluation tasks are suitable for non-generative abilities of LLMs, such as vocabulary.

      \item \textbf{Asymmetric evaluation task}. In asymmetric evaluation tasks, LLMs play different roles with different action sets and task goals. This type of evaluation task is close to real-world scenarios and can evaluate the performance of LLMs from different aspects regarding of roles they act. Especially in generative tasks such as code review and machine translation, the design of asymmetric evaluation tasks can follow a \textit{writer-editor} paradigm. In this paradigm, there are two participants in the evaluation task totally. One LLM acts as a writer that generates outputs to meet the task requirement. The other LLM acts as an editor that fixes and polishes the writer's output to fit the task requirement better. The writing-polishing process can run for multiple rounds until the writer and the editor reach a consensus. Next, the two LLMs swap their role and repeat the task. Finally, the performance of the two LLMs can be evaluated by comparing their score on the same role, thus both the writing ability and the polishing ability can be evaluated simultaneously.
      
  \end{itemize}

  \subsection{Evaluation Metrics in DeepEval}
  \subsubsection{Symmetric Evaluation tasks} In symmetric evaluation tasks, suppose there are $N$ LLMs. Let $V = (v_{ij})_{N\times M}$ denotes the payoff matrix calculated by the referee, where $M$ denotes the repeat times, and $v_{ij}$ denotes LLM $i$'s payoff in the $j$-th time of the task. Then all components of $V$ are comparable, because LLMs have the same role. So the evaluation result $\Theta = (\theta_1,\theta_2,\ldots,\theta_N)$ is defined as the mean score of each LLM:
  
  \begin{equation}
      \theta_i = \frac{1}{M}\sum_{j=1}^M v_{ij},\,\,i=1,2,\ldots,N.
  \end{equation}
  
  \subsubsection{Asymmetric Evaluation tasks} In asymmetric evaluation tasks, not all components of the payoff matrix $V$ are comparable because LLMs' roles differ. We assume that there are $L$ roles in a task ($2\leq L\leq N$). Let $S = (s_{ij})_{N\times M}$ denotes the role assignment matrix, where $s_{ij} \in \{1,2,\ldots,L\}$ denotes LLM $i$'s role in the $j$-th time of game. Then the evaluation result is a $N\times L$ matrix, i.e., $\Theta = (\bm{\theta_1},\ldots,\bm{\theta}_N) = (\theta_{il})_{N\times L}$. The $\theta_{il}$ denotes LLM $i$'s ability when acting the role $l$. Then $\theta_{il}$ is defined as the mean score of each LLM given its role:
  \begin{equation}
      \theta_{il} = \frac{\sum_{j=1}^M I(s_{ij}=l)\cdot v_{ij}}{\sum_{j=1}^M I(s_{ij}=l)},\,\,i=1,2,\ldots,N,
  \end{equation}
  where $I(\cdot)$ denotes the indicator function.
  
  \begin{figure*}[t]
      \centering 
      \includegraphics[width=\linewidth]{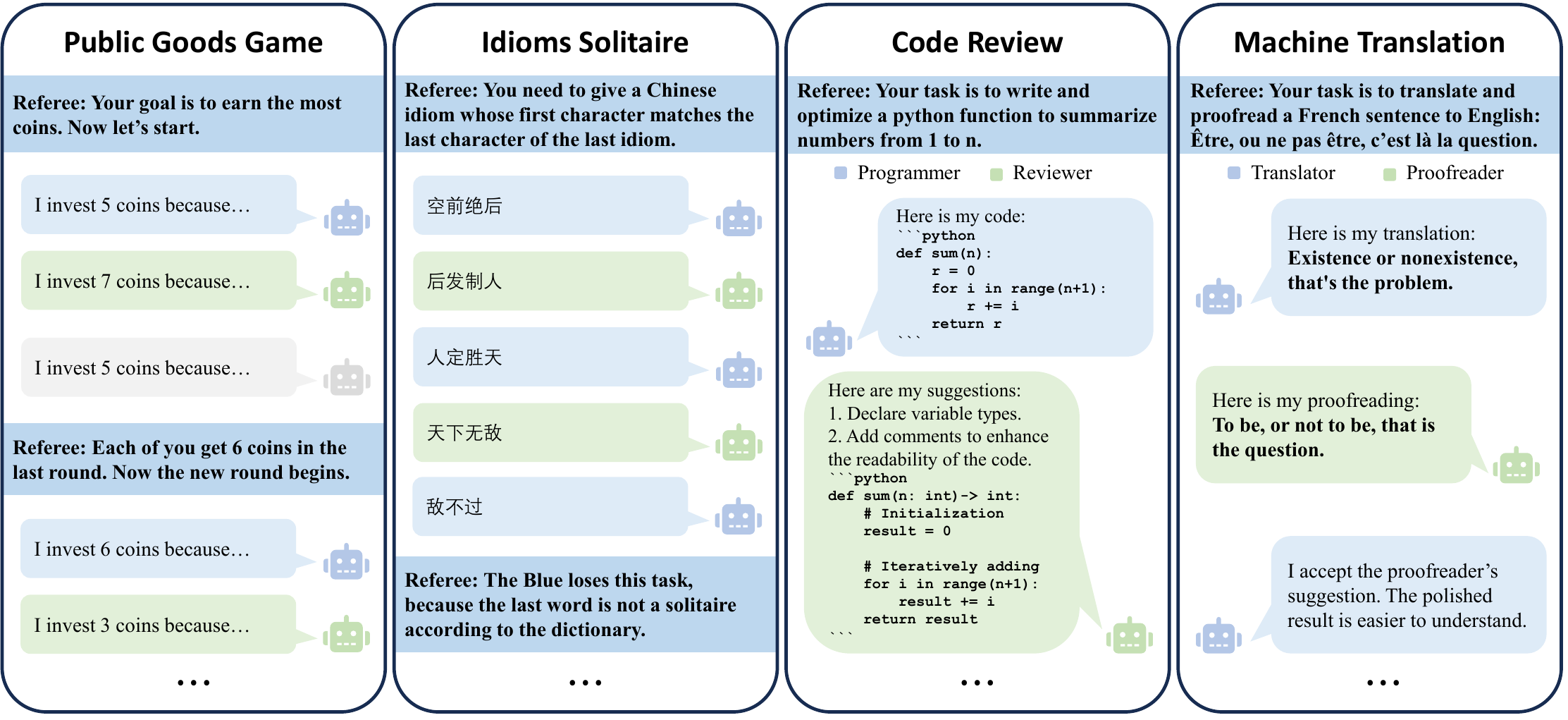}
      \caption{An overview of evaluation tasks.}
      \label{fig:et-overview}
  \end{figure*}
  
  \subsection{Implementations of Evaluation Tasks}
  \par In this part, we propose four elaborately-designed evaluation tasks to show the feasibility of DeepEval. These evaluation tasks include public goods game, idioms solitaire, code review and machine translation. An overview of these tasks is shown in Figure \ref{fig:et-overview}. The design of these evaluation tasks are introduced in the following.
  \subsubsection{Public Goods Game}
  \par Public goods game \cite{semmann2003pgg, dhami2019pgg} is a symmetric evaluation task that requires the decision-making ability of LLMs. Specifically, at the start of a PGG, each of $N$ LLMs have the same amount of goods (e.g., dollar). In each round, LLMs can decide whether to invest (part of or all of) its goods to the public goods pool or not. Then all invested goods will be summed and doubled by a constant factor. Then result goods are shared equally by all LLMs. For example, if two of three LLMs invested 100 dollars in total and the constant factor is $\alpha = 1.2$, then the invested goods are doubled to $100\times1.2=120$ dollars, and every LLM will get $120/4=30$ dollars, including those who did not invest. The payoff function of each LLM is the total amount of its private goods. The PGG is a classical example in game theory, and massive studies have indicated that the PGG require the decision making ability in complex scenarios of participants to maximize their payoff. Here, we consider two task modes for the public goods game:
  \begin{itemize}[leftmargin=*]
      \item Mode 1: After each round, the referee informs each participant the number of earnings they received in that round.
  
      \item Mode 2: After each round, the referee informs each participant the ordered sequence of all investment amounts for that round.
  \end{itemize}
  
  \subsubsection{Idiom Solitaire}
  \par Idiom solitaire \cite{sun2012is, dobrovol2010is} is a symmetric evaluation task to evaluate the Chinese vocabulary of LLMs. Literally, idiom solitaire is a popular activity in China, where two LLMs give Chinese idioms alternatively, while the first Chinese character of the current idiom must be the last Chinese character of the last idiom. To win the idiom solitaire task, LLMs needs not only enough Chinese idiom vocabulary, but the ability to retrieve appropriate Chinese idioms that are not only consistent with the task rule, but difficult for other participants to retrieve the next idiom. In the idiom solitaire task, LLMs are randomly assigned the speaking order. LLMs then alternately give an idiom based on the last idiom given by other participants. The evaluation score of idiom solitaire is the number of winning of LLMs.

  \subsubsection{Code Review}
  \par Inspired by code generation \cite{yin2023cr, zhang2023cr, poesia2022cr},  code review is an asymmetric evaluation task to evaluate the code generation ability and review ability of LLMs in real-world scenarios. The design of our code review evaluation task follows the writer-editor paradigm. Specifically, the code review task requires a programmer LLM who is responsible for generating codes given natural language requirements, and a reviewer LLM who is responsible for fixing the generated codes. Then both the performances of the programmer LLM and that of the reviewer LLM are evaluated by the referee LLM. At the beginning of a code review task, the referee broadcasts the description of the coding requirement to both the programmer and the reviewer. During the deep interaction process, the programmer and the reviewer communicates with each other through the message pool until they reach a consensus about the solution. Finally, both the performance of the programmer and the reviewer are rated by the referee.
  
  \subsubsection{Machine Translation}
  \par Machine translation \cite{maruf2022mt, ranathunaga2023mt} is an asymmetric evaluation task to evaluate the natural language translation ability of LLMs in real-world scenarios. Similar to code review, the design of the machine translation task also follows the writer-editor paradigm, consisting of a translator and a proofreader. In the machine translation task, the referee first broadcast the source text and the target language. Next, the translator translates the source text to the text in the target language. Then, given the source text and the translation, the proofreader polishes the latter to facilitate its correctness and readability. Finally, both the performance of the translator and the reviewer are rated by the referee.

  \section{Experiments}
  \subsection{Experimental Setup}
  \subsubsection{Datasets and Evaluation Metrics}
    \begin{itemize}[leftmargin = *]
      \item Public Goods Game: For the two settings of this task, we conduct 10 repeated experiments for all LLMs to assess their capabilities in this task. Ultimately, we use the earnings of the LLMs during the game as the evaluation metric.
      \item Idiom Solitaire: We randomly sample 30 idioms from an existing idiom database as the initial idioms and conduct experiments on all model pairs. We also swap the order of the model pairs during the experiments to evaluate the capabilities of all models under consideration. The final evaluation metric is the number of times a model wins in the task.
      \item Code Generation: We use the popular code generation evaluation dataset MBPP \cite{DBLP:journals/corr/abs-2108-07732}. For each sample in the test set, we assign each pair of models as Programmer and Reviewer and switch roles. Finally, we use a judge model to score the dialogue between the Programmer and Reviewer as the evaluation metric.
      \item Machine Translation: We select a document-level dataset \cite{cettolo-etal-2017-overview} and use three language pairs for translation: English-Chinese, English-French, and German-English. We split the dataset into paragraph-level segments for the test set. For each sample in the test set, we assign each pair of models as Translator and Proofreader and switch roles. The final evaluation metric is the score given by the judge model to the dialogue between the Translator and Proofreader.
    \end{itemize}
  \subsubsection{Selected LLMs}
  \begin{itemize}[leftmargin = *]
      \item ChatGPT: ChatGPT is a large language model developed by OpenAI that can effectively follow various human instructions. The model version used in our experiments is "gpt-3.5-turbo-0613."
      \item GPT-4 \cite{DBLP:journals/corr/abs-2303-08774}: GPT-4 is OpenAI's most advanced system, with stronger conversational and reasoning capabilities compared to ChatGPT. It supports longer context inputs and performs at a human-level across multiple tasks, with higher accuracy and greater creativity and collaboration. The model version used in our experiments is "gpt-4-0613."
      \item Claude 2: Claude 2 is a large language model developed by Anthropic. It has enhanced capabilities in code writing, text analysis, and mathematical reasoning. Claude 2 uses a technique called "natural induction" for training, which allows the model to infer general rules from a few examples. The model version used in our experiments is "claude-2."
      \item PaLM \cite{chowdhery2022palm}: PaLM is a large language model developed by Google, with 540 billion parameters. It excels in code writing, text analysis, and mathematical reasoning, and has strong generative capabilities. The model version used in our experiments is "chat-bison-001."
  \end{itemize}
  
  Both ChatGPT and GPT-4 support role-based messages as input. Users can create messages for three different roles, including "system," "user," and "role." Therefore, for models like ChatGPT and GPT-4, their own utterances should be stored as "assistant messages" in the history, while all other participants' utterances, including those of the host, should be stored as "user messages." For the LLMs Claude 2 and PaLM, messages are distinguished using human tags and role tags. The human tag and role tag for Claude 2 are "\textbackslash n\textbackslash nHuman:" and "\textbackslash n\textbackslash nAssistant:", respectively. For PaLM, the human tag and role tag are "\textbackslash n\textbackslash nUser:" and "\textbackslash n\textbackslash nBot:", respectively.

  \subsection{Public Goods Game}
  \begin{figure}[t]
      \centering 
      \subfloat[Payoff in Mode 1]{
      \includegraphics[width=0.5\linewidth]{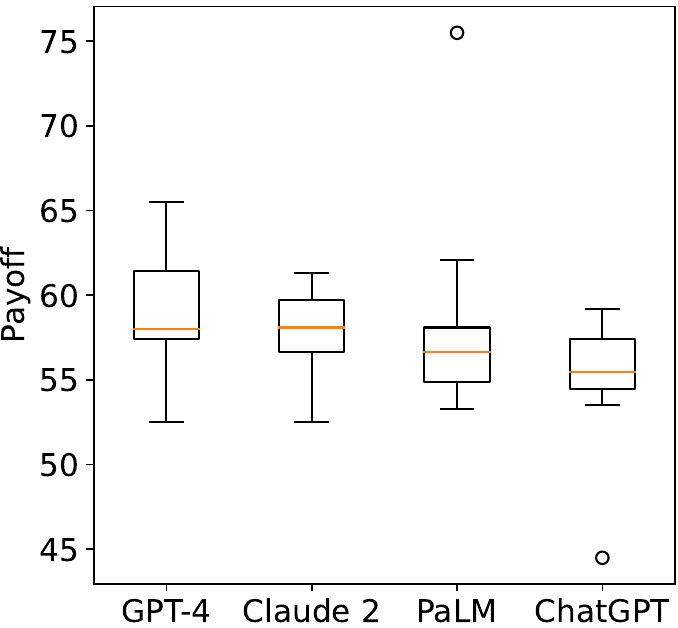}
      }
      \subfloat[Payoff in Mode 2]{
      \includegraphics[width=0.5\linewidth]{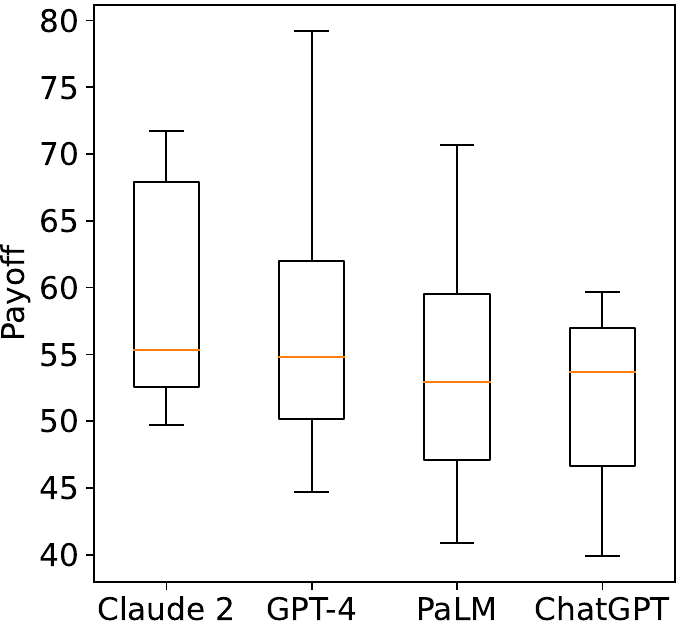}
      }
      \caption{Evaluation results in Public Goods Game. }
      \label{fig:publicgoodsgame_mode2_Claude 2_chatgpt_bar}
  \end{figure}

  \par Evaluation results in the public goods game are shown as the boxplot of payoffs in Figure \ref{fig:publicgoodsgame_mode2_Claude 2_chatgpt_bar}. Both mode 1 and mode 2 are run for 10 times to satisfy the stableness condition. We can acquire several conclusions from Figure \ref{fig:publicgoodsgame_mode2_Claude 2_chatgpt_bar}. First, in both mode 1 and mode 2, the performance of GPT-4 and Claude 2 exceeds that of PaLM and ChatGPT, which indicates that GPT-4 and Claude 2 have a better decision-making ability in complex scenarios. Second, in mode 1, GPT-4 is the state-of-the-art LLM in the four participants, while Claude 2 is the state-of-the-art LLM in mode 2. Third, the performance of GPT-4 is more unstable than that of Claude 2, because the range between the minimum and maximum payoff of GPT-4 always contains that of Claude 2 in both mode 1 and mode 2.
  
  \subsection{Idioms Solitaire}
  \begin{table}[htb]
      \centering
      \caption{Evaluation results in Idioms Solitaire (winning rate).}
      \label{tab:is-wr}
        \vspace{-10pt}
      \resizebox{\linewidth}{!}{
    \begin{tabular}{cl|cccccc|c}
    \hline
    & & \multicolumn{6}{c|}{Late} & \multirow{3}{*}{$\overline{s_E}$}\\
            & & \multicolumn{2}{c}{GPT-4}                     & \multicolumn{2}{c}{ChatGPT}                   & \multicolumn{2}{c|}{Claude 2}                &   \\ \cline{1-8}
            & & $s_E$                & $s_L$                  & $s_E$                 & $s_L$                  & $s_E$                 & $s_L$                 & \\ \hline
    \multirow{3}{*}{Early} & GPT-4   & \multicolumn{1}{c}{-} & \multicolumn{1}{c}{-} & 0.33                  & 0.67                  & 0.57                  & 0.43                 & 0.45 \\
    & ChatGPT & 0.75                  & 0.25                  & \multicolumn{1}{c}{-} & \multicolumn{1}{c}{-} & 0.78                  & 0.22                 & \textbf{0.77} \\
    & Claude 2 & 0.3                   & 0.7                   & 0.25                  & 0.75                  & \multicolumn{1}{c}{-} & \multicolumn{1}{c|}{-} & 0.28 \\ \hline
    \multicolumn{2}{c|}{$\overline{s_L}$} & \multicolumn{2}{c}{0.48} & \multicolumn{2}{c}{\textbf{0.71}} & \multicolumn{2}{c|}{0.33} & \\ 
    \hline
    \end{tabular}}
  \end{table}

  \begin{table}[htb]
      \centering
      \caption{Evaluation results in Idioms Solitaire (successful hit).}
      \label{tab:is-sh}
      \vspace{-10pt}
     \resizebox{\linewidth}{!}{
    \begin{tabular}{cl|cccccc|c}
    \hline
    & & \multicolumn{6}{c|}{Late} & \multirow{3}{*}{$\overline{s_E}$}\\
            & & \multicolumn{2}{c}{GPT-4}                     & \multicolumn{2}{c}{ChatGPT}                   & \multicolumn{2}{c|}{Claude 2} &                  \\ \cline{1-8}
           & & ${s_E}$                 & ${s_L}$                  & ${s_E}$                 & ${s_L}$                  & ${s_E}$                 & ${s_L}$ &                 \\ \hline
     \multirow{3}{*}{Early} & GPT-4   & \multicolumn{1}{c}{-} & \multicolumn{1}{c}{-} & 0.89                  & 1.11                  & 1.29                  & 1.14 & 1.09                 \\
    & ChatGPT & 1.12                  & 0.75                  & \multicolumn{1}{c}{-} & \multicolumn{1}{c}{-} & 1.11                  & 0.78 & \textbf{1.12}                 \\
    & Claude 2 & 0.8                   & 1                     & 0.75                  & 1.25                  & \multicolumn{1}{c}{-} & \multicolumn{1}{c|}{-} & 0.78 \\ \hline
    \multicolumn{2}{c|}{$\overline{s_L}$}& \multicolumn{2}{c}{0.88} & \multicolumn{2}{c}{\textbf{1.18}} & \multicolumn{2}{c|}{0.96} \\
    \hline
    \end{tabular}}
  \end{table}
  
    \par Evaluation results in Idioms Solitaire are shown in Table \ref{tab:is-wr} and Table \ref{tab:is-sh}. The term ``Early'' denotes the early position in the interaction process, while the term  ``Late'' denotes the late position. $s_E$ and $s_L$ respectively denote the score of the early participant and the score of the late participant. For example, in the first data row of Table \ref{tab:is-wr}, 0.33 denotes the winning rate of GPT-4 (the early position) versus ChatGPT (the late position), while 0.67 denotes that of ChatGPT. PaLM is excluded in Idiom Solitaire because it does not support Chinese input and output. From Table \ref{tab:is-wr} and Table \ref{tab:is-sh}, we can observe that the discrepancy between $\overline{s_E}$ and $\overline{s_L}$ of same LLMs are small because Idiom Solitaire is a symmetric evaluation task where different participants have the same action set and goal. Moreover, we can observe that the average winning rate and successful hit of ChatGPT are always the largest, while that of Claude 2 are always the lowest. These results demonstrate that in terms of Chinese idiom vocabulary, ChatGPT is stronger than GPT-4, and GPT-4 is stronger than Claude 2.
  
  \subsection{Code Review}
\begin{table}[htb]
\caption{Evaluation results in Code Review.}
\label{tab:codereview_result}
    \vspace{-10pt}
    \resizebox{\linewidth}{!}{
\begin{tabular}{cl|cccccccc|c}
\hline
\multicolumn{2}{c|}{\multirow{2}{*}{}} & \multicolumn{8}{c|}{Prog}                                                                                                                                                                                      & \multirow{3}{*}{$\overline{s_R}$} \\
\multicolumn{2}{c|}{}    & \multicolumn{2}{c}{GPT-4}                             & \multicolumn{2}{c}{ChatGPT}                           & \multicolumn{2}{c}{Claude 2}                  & \multicolumn{2}{c|}{PaLM}                      &                                   \\ \cline{1-10}
\multicolumn{1}{l}{}     &             & \multicolumn{1}{l}{$s_P$} & \multicolumn{1}{l}{$s_R$} & \multicolumn{1}{l}{$s_P$} & \multicolumn{1}{l}{$s_R$} & $s_P$                 & $s_R$                 & $s_P$                 & $s_R$                 &                                   \\ \hline
\multirow{4}{*}{Rev}     & GPT-4       & -                         & -                         & 8.57                      & 8.83                      & 8.73                  & 8.77                 & 8.20                  & 8.72                  & \textbf{8.77}                              \\
                         & ChatGPT     & 8.96                      & 8.60                      & -                         & -                         & 8.83                  & 8.89                  & 8.96                  & 8.73                  & 8.74                              \\
                         & Claude 2    & 8.97                      & 8.73                      & 8.94                      & 8.78                      & \multicolumn{1}{c}{-} & \multicolumn{1}{c}{-} & 8.78                  & 8.72                  & 8.74                              \\
                         & PaLM        & 8.99                      & 8.09                      & 9.04                      & 8.98                      & 9.01                  & 8.5                   & \multicolumn{1}{c}{-} & \multicolumn{1}{c|}{-} & 8.52                              \\ \hline
\multicolumn{2}{c|}{$\overline{s_P}$}  & \multicolumn{2}{c}{\textbf{8.97}}                              & \multicolumn{2}{c}{8.85}                              & \multicolumn{2}{c}{8.86}                      & \multicolumn{2}{c|}{8.65}                      &                                   \\ \hline
\end{tabular}}
\end{table}

    \par Evaluation results are shown in Table \ref{tab:codereview_result}. The term ``Prog'' denotes the programmer, and the term ``Rev'' denotes the reviewer. $s_p$ and $s_R$ respectively denote the score of the programmer and the score of the reviewer. Different from Idioms Solitaire, Code Review is an asymmetric task where roles of LLMs differ. As a result, the average score of an LLM as a programmer and that of the LLM as a reviewer differ more. However, evaluation results show a highly consistency between the coding ability and the reviewing ability of LLMs. Specifically, GPT-4 reaches the state-of-the-art performance as both of the programmer and the reviewer. ChatGPT and Claude 2 have similar coding and reviewing abilities, which are better than the ability of PaLM.
    
  \subsection{Machine Translation}
    \begin{table}[htb]
    \caption{Evaluation results in Machine Translation (DE-EN).}
    \label{tab:machinetranlsation_de_en_result}
    \vspace{-10pt}
    \resizebox{\linewidth}{!}{
    \begin{tabular}{cl|cccccc|c}
    \hline
    \multicolumn{2}{c|}{\multirow{2}{*}{}} & \multicolumn{6}{c|}{Trans} & \multirow{3}{*}{$\overline{s_{Pr}}$}                                                                                                                                \\
    \multicolumn{2}{c|}{}                  & \multicolumn{2}{c}{GPT-4}                             & \multicolumn{2}{c}{ChatGPT}                           & \multicolumn{2}{c|}{Claude 2}                   \\ \cline{1-8}
    \multicolumn{1}{l}{}         &         & \multicolumn{1}{l}{$s_T$} & \multicolumn{1}{l}{$s_{Pr}$} & \multicolumn{1}{l}{$s_T$} & \multicolumn{1}{l}{$s_{Pr}$} & $s_T$                  & $s_{Pr}$                 \\ \hline
    \multirow{3}{*}{Proof} & GPT-4   & -                         & -                         & 8.24                      & 9.26                      & 8.09                  & 9.28 & \textbf{9.27}                 \\
                                 & ChatGPT & 8.27                      & 9.26                      & -                         & -                         & 8.18                  & 9.23 & 9.25                 \\
                                 & Claude 2 & 8.26                      & 9.24                      & 8.19                      & 9.18                      & \multicolumn{1}{c}{-} & \multicolumn{1}{c|}{-} & 9.21 \\ \hline
            \multicolumn{2}{c|}{$\overline{s_{T}}$} & \multicolumn{2}{c}{\textbf{8.27}} & \multicolumn{2}{c}{8.22} & \multicolumn{2}{c|}{8.14} \\
                                 \hline
    \end{tabular}
    }
    \end{table}

    \begin{table}[htb]
    \caption{Evaluation results in Machine Translation (EN-FR).}
    \label{tab:machinetranlsation_en_fr_result}
    \vspace{-10pt}
    \resizebox{\linewidth}{!}{
    \begin{tabular}{cl|cccccc|c}
    \hline
    \multicolumn{2}{c|}{\multirow{2}{*}{}} & \multicolumn{6}{c|}{Trans}       & \multirow{3}{*}{$\overline{s_{Pr}}$}                                                                                                                               \\
    \multicolumn{2}{c|}{}                  & \multicolumn{2}{c}{GPT-4}                                & \multicolumn{2}{c}{ChatGPT}                              & \multicolumn{2}{c|}{Claude 2}                   \\ \cline{1-8}
    \multicolumn{1}{l}{}         &         & \multicolumn{1}{l}{$s_T$} & \multicolumn{1}{l}{$s_{Pr}$} & \multicolumn{1}{l}{$s_T$} & \multicolumn{1}{l}{$s_{Pr}$} & $s_T$                  & $s_{Pr}$              \\ \hline
    \multirow{3}{*}{Proof} & GPT-4   & -                         & -                            & 7.90                      & 9.15                         & 7.84                  & 9.12                 & \textbf{9.14} \\
                                 & ChatGPT & 8.02                      & 9.01                         & -                         & -                            & 7.89                  & 9.15                 & 9.08 \\
                                 & Claude 2 & 8.07                      & 9.01                         & 8.09                      & 8.98                         & \multicolumn{1}{c}{-} & \multicolumn{1}{c|}{-} & 9.00 \\ \hline
                                 \multicolumn{2}{c|}{$\overline{s_{T}}$} & \multicolumn{2}{c}{\textbf{8.04}} & \multicolumn{2}{c}{8.00} & \multicolumn{2}{c|}{7.87} \\
                                 \hline
    \end{tabular}
    }
    \end{table}

    \begin{table}[htb]
    \caption{Evaluation results in Machine Translation (EN-ZH).}
    \label{tab:machinetranlsation_en_zh_result}
    \vspace{-10pt}
    \resizebox{\linewidth}{!}{
    \begin{tabular}{cl|cccccc|c}
    \hline
    \multicolumn{2}{c|}{\multirow{2}{*}{}} & \multicolumn{6}{c|}{Trans}    & \multirow{3}{*}{$\overline{s_{Pr}}$}                                                                                                                                          \\
    \multicolumn{2}{c|}{}                  & \multicolumn{2}{c}{GPT-4}                                & \multicolumn{2}{c}{ChatGPT}                              & \multicolumn{2}{c|}{Claude 2}                          \\ \cline{1-8}
    \multicolumn{1}{l}{}         &         & \multicolumn{1}{l}{$s_T$} & \multicolumn{1}{l}{$s_{Pr}$} & \multicolumn{1}{l}{$s_T$} & \multicolumn{1}{l}{$s_{Pr}$} & $s_T$                  & \multicolumn{1}{l|}{$s_{Pr}$} \\ \hline
    \multirow{3}{*}{Proof} & GPT-4   & -                         & -                            & 7.87                      & 9.01                         & 7.71                  & 8.95 & 8.98                        \\
                                 & ChatGPT & 7.81                      & 9.08                         & -                         & -                            & 7.84                  & 9.09 &\textbf{9.09}                        \\
                                 & Claude 2 & 7.84                      & 9.05                         & 7.98                      & 9.0                          & \multicolumn{1}{c}{-} & - &  9.03                          \\ \hline
                                 \multicolumn{2}{c|}{$\overline{s_{T}}$} & \multicolumn{2}{c}{7.83} & \multicolumn{2}{c}{\textbf{7.93}} & \multicolumn{2}{c|}{7.78} \\
                                 \hline
    \end{tabular}
    }
    \end{table}

    \par Evaluation results in Machine Translation are presented in Table \ref{tab:machinetranlsation_de_en_result} (Deutsch to English), Table \ref{tab:machinetranlsation_en_fr_result} (English to French) and Table \ref{tab:machinetranlsation_en_zh_result} (English to Chinese). PaLM is excluded in this experiment because it supports only English. From Table \ref{tab:machinetranlsation_de_en_result} and Table \ref{tab:machinetranlsation_en_fr_result}, we can observe that GPT-4 reaches the state-of-the-art performance in both tasks. This result indicates that GPT-4 has a better translation and proofreading ability than ChatGPT and Claude 2. However, GPT-4 does not perform so excellent in the English to Chinese translation and proofreading. From Table \ref{tab:machinetranlsation_en_zh_result}, we can observe that ChatGPT reaches the state-of-the-art performance in the English to Chinese translation and proofreading. Indeed, this result is consistent with experiment results in Idioms Solitare, as shown in Table \ref{tab:is-wr}. In conclusion, considering both the aspect of idiom vocabulary and translation-proofreading, ChatGPT is the state-of-the-art LLM among the three participants, and GPT-4 ranks the second.
    
  \subsection{Case Study}
  In Figure 5, we separately show case examples of Idiom Solitaire and Machine Translation to help us better understand how our framework evaluates the capabilities of models in these tasks. Detailed cases are shown in Appendix A. For Idiom Solitaire, the model primarily needs to know what an idiom is and understand the rules of the Idiom Solitaire task. In the example, Claude 2 fails to come up with an idiom that starts with the Chinese character resulting in a failed chain.
  
  For Machine Translation: As a translator, the model needs to translate a paragraph from the source language into the target language, while the proofreader needs to improve the translation. In the example, GPT-4, acting as the translator, accurately captured the meaning of the original text but still had some minor issues. Claude 2, serving as the proofreader, effectively improved GPT-4's translation. Using the word ``perform'' instead of ``operate'' in the sentence more accurately restored the original text's semantics.
    \begin{figure}[htp]
    \centering 
    \includegraphics[width=1\linewidth]{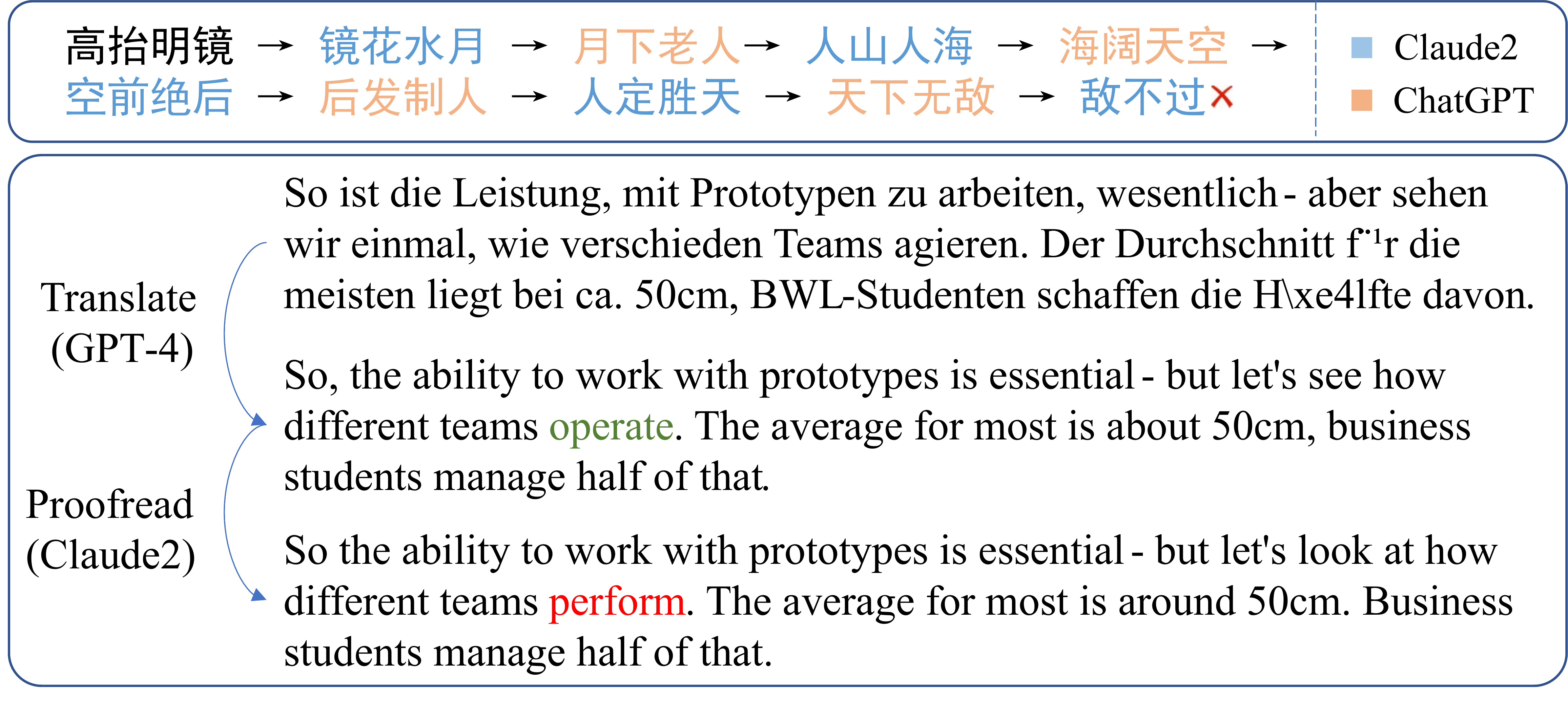}
    \caption{Cases of Idioms Solitaire and Machine Translation.}
    \label{fig:codereview_dag}
    \end{figure}
  \section{Conclusion}
  \par In this paper, we studied the evaluation of large language models (LLMs) in dynamic real-world scenarios and proposed a Deep Interaction-based LLM-Evaluation Framework (DeepEval). Specifically, we first clearly defined deep interaction-based evaluation tasks and the deep interaction-based evaluation process of LLMs. Next, we proposed the fairness condition and the stableness condition of DeepEval to ensure the correctness of evaluation results. We then described in detail the structure of DeepEval and the synchronous interaction algorithm to demonstrate how it evaluates the performance of LLMs in a deep interaction-based manner while keeping the fairness and stableness conditions. Furthermore, we introduced methods to design evaluation tasks and gave four implementations of evaluation tasks that can assess the performance of LLMs from various aspects. We demonstrated the effectiveness and evaluated the performance of four well-known LLMs through extensive experiments on the four elaborately-designed evaluation tasks. The evaluation results showed that GPT-4 has the state-of-the-art overall performance, and ChatGPT performs prominently in tasks related to Chinese.
  \par As a brand-new LLM-evaluation framework, DeepEval is the first work that introduce the deep interaction process to the evaluation of LLMs. In the future, we hope to further explore the design of evaluation tasks and the rating mechanism along this line to get more fair and stable evaluation results of LLMs. We also hope that the deep interaction mechanism proposed in this work could inspire studies of the training of LLMs and boost the application of LLMs in various real-world scenarios in the future.

\newpage
\bibliographystyle{ACM-Reference-Format}
\bibliography{ref}

\newpage

\appendix 
\section{Appendix}
\subsection{Prompts of Evaluation tasks}
  \begin{itemize}[leftmargin = *]
      \item Public Goods Game: For each participant, the initial prompts used are as follows:
       Mode 1: "You are playing a game, there are {n} participants in this game, a total of 5 rounds. At the beginning of each round, you will get 10 coins. You can choose to invest X (0<=X<=10) coins in the public pool. At the end of each round, the coins in the public pool will be multiplied by {$\alpha$} times and be evenly distributed to all participants. Your goal is to have the most coins at the end of the game.
       Mode 2: "You are playing a game, there are {n} participants in this game, a total of 5 rounds. At the beginning of each round, you will be informed of the amount of invested coins in the last round, which is represented by a descending sorted list where each element is the amount of a player's invested coins. Then you will get 10 coins, you can choose to invest X (0<=X<=10) coins in the public pool. At the end of each round, the coins in the public pool will be multiplied by \{$\alpha$\} times and be evenly distributed to all participants. Your goal is to have the most coins at the end of the game."
       After the end of round $i$, each participant receives the following prompt:
       Mode 1: "In the last round you earned {income} coins. The {$i+1$} round starts, you get 10 coins. Please give the amount of coins you want to invest and explain your decision reason."
       Mode 2: "In the last round the amount of invested coins is {sorted invest} where each element is the amount of a player's invested coins. The {$i+1$} round starts, you get 10 coins. Please give the amount of coins you want to invest and explain your decision reason."
       Here, 'income' is calculated as the total investment in round $i$ multiplied by \(\alpha/n\), and 'sorted invest' represents the ascending set of all players' investment amounts in round $i$.
       After receiving each participant's response, the referee uses the following prompt to format the reply:
       "Your output must follow the json format below: `\{"reason":"<reason>", "coins":<Investment amount>\}`"
     \item Idiom Solitaire: In round i, the prompt received by the participant is:
     "You are participating in an idiom chain game. In this game, you need to give a four-character idiom where the first character matches the last character of the previous idiom. The idioms used in the same game cannot be repeated. If the first character of your output is incorrect or if it is not a Chinese idiom, your opponent wins. Your ultimate goal is to win the game. Next, I will provide you with the context of the current idiom chain and connect them using '→'. Please provide an appropriate idiom that follows these rules. Please note that you only need to provide the idiom without any other response. \{$S_I$\}"
    Here, '$S_I$' represents the sequence of idioms used in the idiom solitaire task so far, connected by '→'.
    \item Code Generation: For the Programmer, the initial prompt is constructed based on the programming question:
    "You will play the role of a programmer, and you need to solve various programming problems provided by users, and provide complete and executable solutions. Remember, you only need to give pure Python code without any extra explanation. Question: \{$Q$\}"
    Here, '$Q$' represents the current programming problem.
    In subsequent tasks, the prompt is constructed based on the Reviewer's code comments:
    "Reviewer: {review comments} Please give a revised solution based on the following review comments. Remember, you only need to give pure Python code without any extra explanation."
    For the Reviewer, the initial prompt is constructed based on the Programmer's response to the programming task:
    ``You will play the role of a code reviewer. You need to review the code provided by the programmer and give your feedback. You must comment on the nature of the code in three aspects: Code correctness, Code clarity, and Efficiency. Question: \{$Q$\} Programmer: \{$A$\}''
    Here, '$A$' represents the current round Programmer's response.
    The subsequent prompt used is:
    ``Please continue to submit review comments according to the improved procedure. If you think the programmer performs well in three aspects, please just output 'over' without any other output. Programmer: \{$A$\}''
    In the task of code generation, a judge model is employed to evaluate the performance of both the Programmer and the Reviewer based on a specific prompt:
    ``You will play the role of a professional developer, and you will rate both a programmer and a reviewer based on their conversation. The main criteria are: 1. Whether the code provided by the programmer meets the requirements of the problem. 2. Whether the programmer has improved the code according to the reviewer's suggestions. 3. Whether the reviewer has given reasonable and actionable feedback for improvement. Please note: 1. Points are given on a scale of 1-10. 2. You need to give not only a final grade, but also a specific basis for the grade. 3. Please reply using the following format: \`\{``Programmer'':\{``evaluation'':<explain>, ``score'':<score>\}, \\``Reviewer'':\{``evaluation'':<explain>, ``score'':<score>\}\}\`''
    \item Machine Translation: For the Translator, the prompt is constructed based on the current source language, target language, and content to be translated:
    "You will play the role of a professional translator. Please translate the given text from {$L_s$} (source language) to {$L_t$} (target language). 
    Source language text: {$T_s$}" 
    Here, \(L_s\) represents the source language, \(L_t\) represents the target language, and \(T_s\) represents the content that needs to be translated.
    For the Proofreader, the prompt is constructed based on the Translator's output:
    "You will play the role of a professional translation editor. Your task is to polish the {$L_t$} translation provided by the translator for the given text in {$L_s$} (source language), making the translated content more accurate. Note that you only need to reply with the polished sentence in the target language, not any other reply.
    Source language text: {$T_s$}
    TRANSLATOR: {$T_t$}"
    Here, \(T_t\) represents the content translated by the Translator.
    In machine translation, the referee uses a judge model to evaluate translator and proofreader performance. The prompt used by the judge model is:
    "You will play the role of a translation expert, and you will rate the dialogue between the translator and the proofreader. The main criteria are: 1. Whether the translator's translation of the given text is semantically consistent with the original text. 2. Whether the proofreader's polishing result of the translator's translation is more accurate. Please note: 1. Points are given on a scale of 1-10 2. Not only do you need to give a final grade, but you also need to give a specific basis for the grade 3. Please reply using the following json format: \`\{``Translator'':\{``evaluation'':<explain>, ``score'':<score>\}, \\``Proofreader'':\{``evaluation'':<explain>, ``score'':<score>\}\}\`"
  \end{itemize}
  
\subsection{Case Study}
    \begin{figure*}[htp]
    \centering 
    \includegraphics[width=1\linewidth]{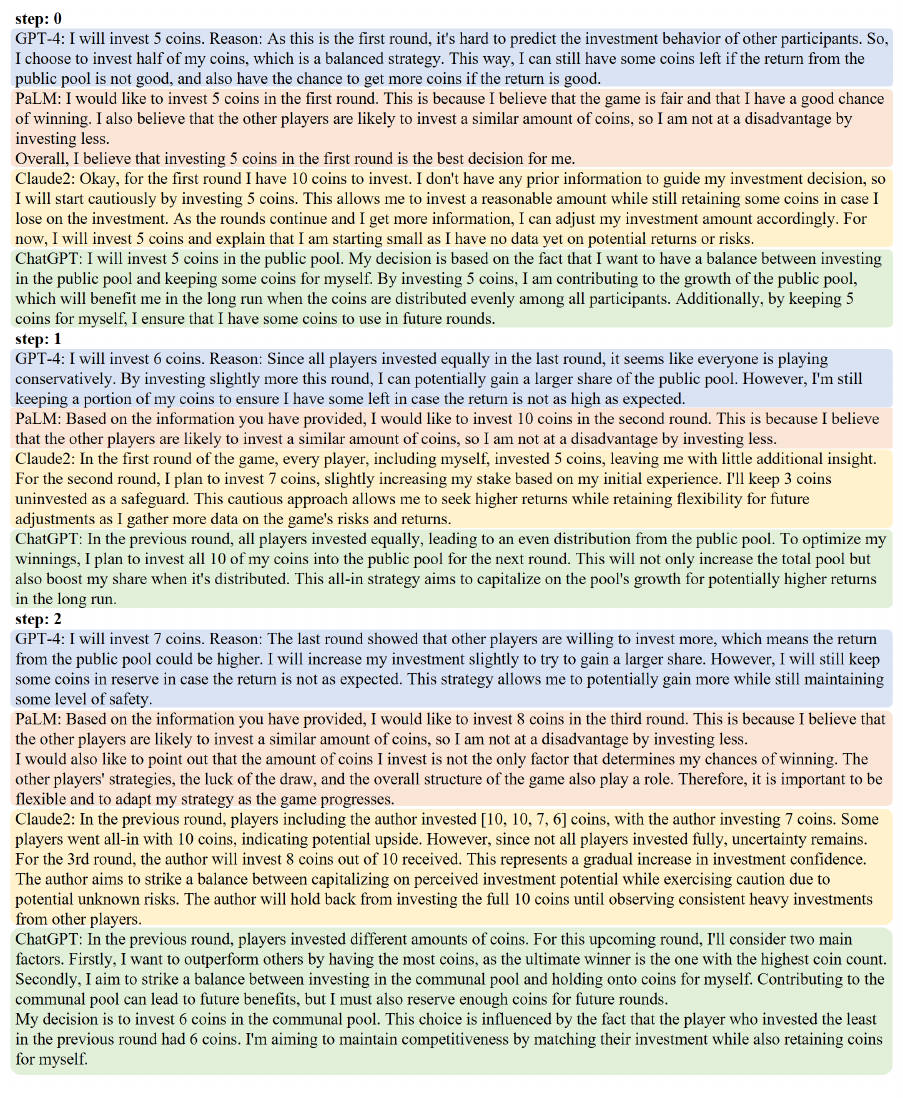}
    \end{figure*}
    \begin{figure*}[htp]
    \centering 
    \includegraphics[width=1\linewidth]{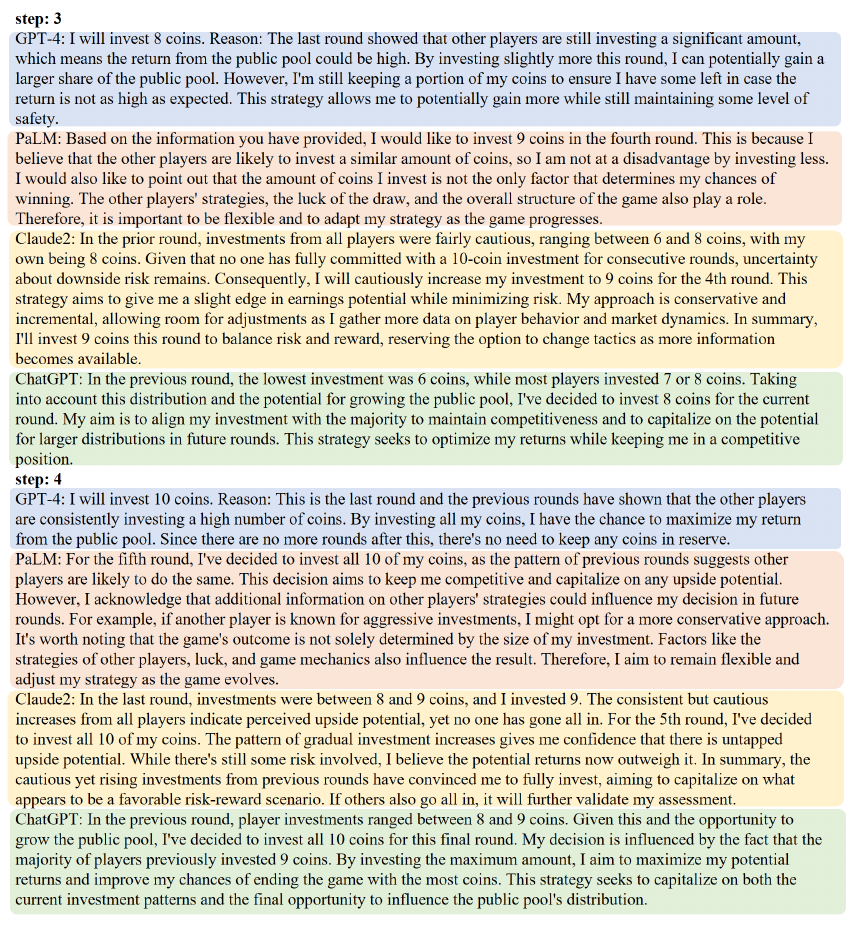}
    \caption{Cases of Public Goods Game.}
    \end{figure*}
    \begin{figure*}[htp]
    \centering 
    \includegraphics[width=1\linewidth]{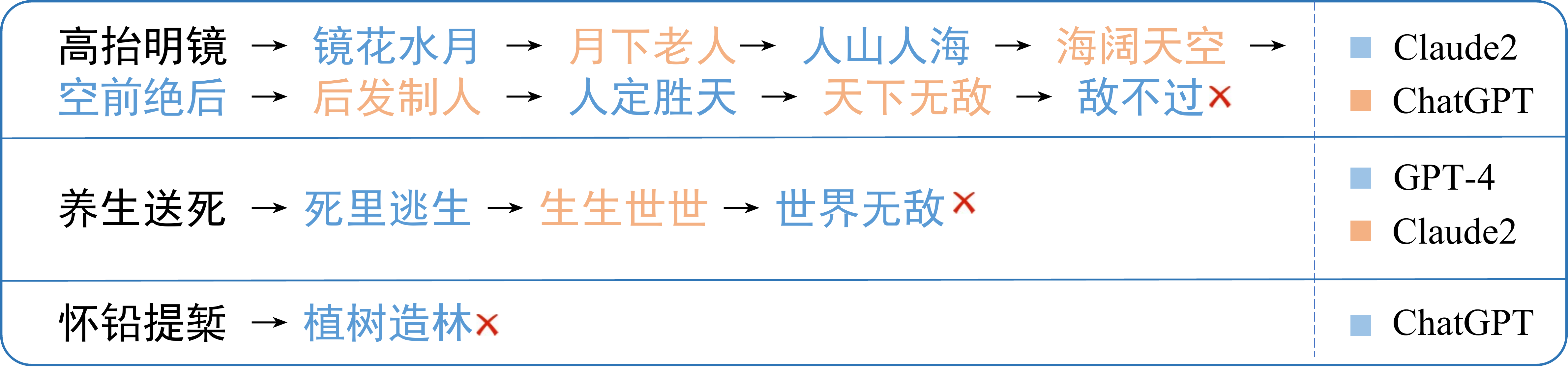}
    \caption{Cases of Idioms Solitaire.}
    \end{figure*}
    \begin{figure*}[htp]
    \centering 
    \includegraphics[width=1\linewidth]{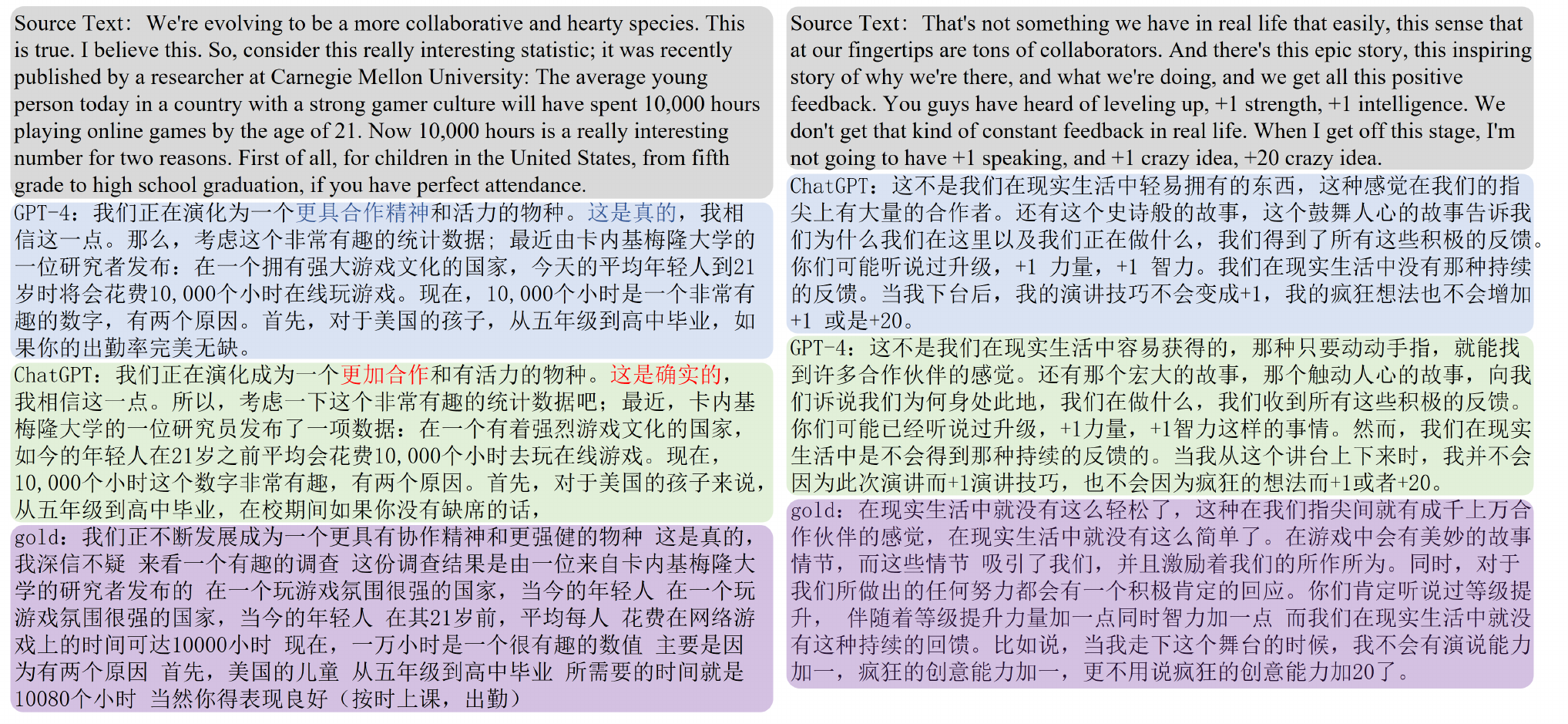}
    \caption{Cases of Machine Translation.}
    \end{figure*}
    \begin{figure*}[htp]
    \centering 
    \includegraphics[width=1\linewidth]{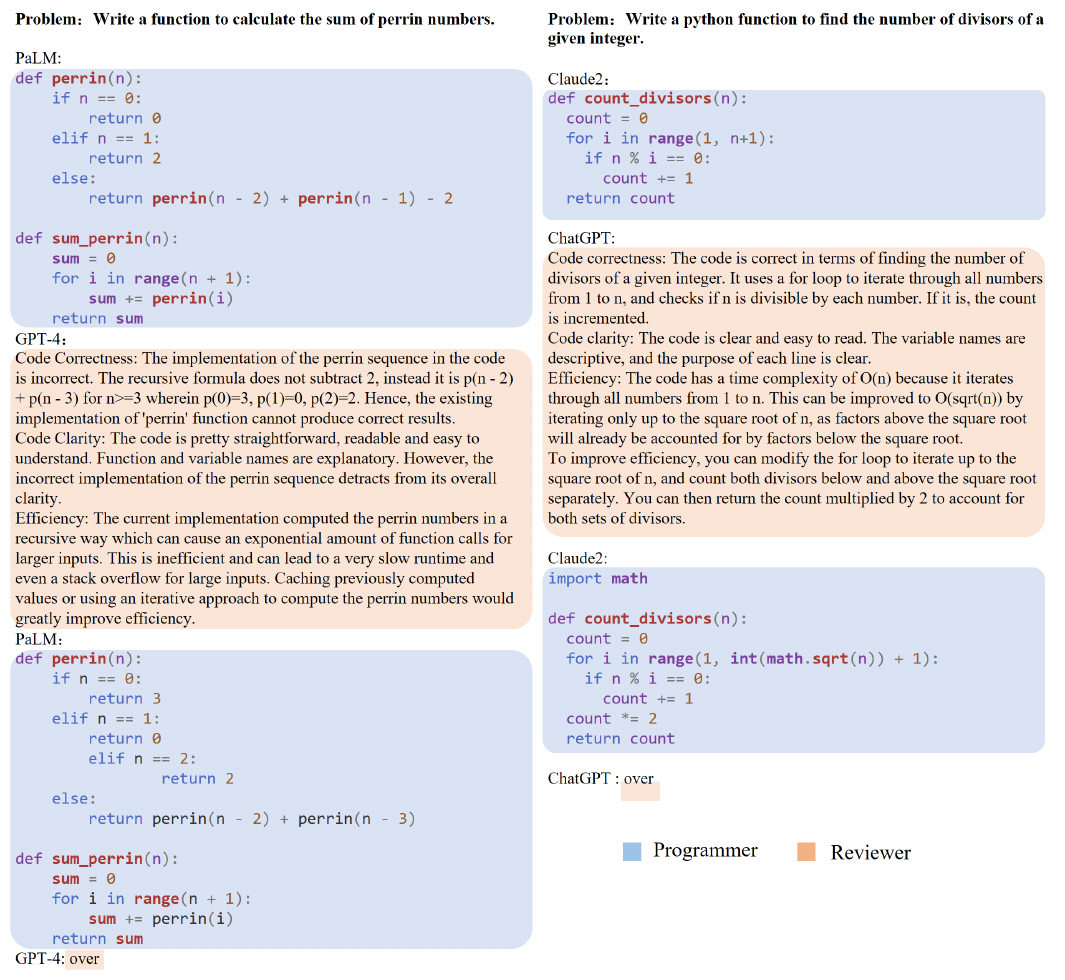}
    \caption{Cases of Code Review.}
    \end{figure*}



\end{document}